\documentclass[twocolumn,apacite,authoryear]{els-mrw} 

\usepackage{amsmath,amssymb,amsfonts,amsthm,makeidx,graphicx}
\usepackage{txfonts}
\usepackage{helvet}
\usepackage[ukrainian,english]{babel}
\usepackage[capitalize]{cleveref}

\DeclareRobustCommand{\cyrins}[1]{%
  \begingroup\fontfamily{erewhon-TLF}%
  \foreignlanguage{ukrainian}{#1}%
  \endgroup
}

\usepackage[normalem]{ulem}
\usepackage{etoolbox}

\begin{document}

\chapter{Formal Constraints on Dependency Syntax}\label{chap1}

\author[1]{Carlos Gómez-Rodríguez}%
\author[2]{Lluís Alemany-Puig}%

\address[1]{\orgname{Universidade da Coruña}, \orgdiv{CITIC, Departamento de Ciencias de la Computación y Tecnologías de la Información}, \orgaddress{Campus de Elviña, s/n, 15071 A Coruña, Spain}}
\address[2]{\orgname{Universitat Politècnica de Catalunya}, \orgdiv{Departament de Ciències de la Computació}, \orgaddress{Campus Nord, 1--3, 08034 Barcelona, Spain}}

\maketitle

\begin{abstract}[Abstract]

Dependency syntax represents the structure of a sentence as a tree composed of dependencies, i.e., directed relations between lexical units. While in its more general form any such tree is allowed, in practice many are not plausible or are very infrequent in attested language. This has motivated a search for constraints characterizing subsets of trees that better fit real linguistic phenomena, providing a more accurate linguistic description, faster parsing or insights on language evolution and human processing. Projectivity is the most well-studied such constraint, but it has been shown to be too restrictive to represent some linguistic phenomena, especially in flexible-word-order languages. Thus, a variety of constraints have been proposed to seek a realistic middle ground between the limitations of projectivity and the excessive leniency of unrestricted dependency structures.


\end{abstract}

\begin{BoxTypeA}[boxlabel]{Key points}
\begin{itemize}
\item Dependency syntax requires formal constraints for descriptive accuracy and efficient parsing.
\item The traditional constraint, projectivity, simplifies parsing but is too restrictive for many natural languages.
\item Mildly non-projective formalisms trade parsing complexity for better linguistic coverage.
\item No single constraint hierarchy captures syntax. Different languages favor different mildly non-projective classes, thus demanding equally diverse parsers.
\end{itemize}
\end{BoxTypeA}

\section{Introduction}\label{chap1:sec1}

Dependency grammar is an approach to syntax where the structure of a sentence is represented by binary, directed relations (called \emph{dependencies}) between its words, which are usually assumed to form a tree. 

Dependency syntax has gained popularity in recent decades both in linguistics and in natural language processing (NLP) \citep{marneffe19}, becoming the dominant approach over its traditional alternative (constituency syntax) in linguistic description of many languages as well as in NLP applications as a whole. The commonly-cited reasons for this increased interest are that, compared to constituency syntax, it offers greater simplicity (as there is a one-to-one correspondence between tokens\footnote{Tokens are the lexical units into which sentences are typically split for syntactic analysis, which include words and punctuation symbols. In what follows, we will loosely use the term ``words'' to refer to tokens.} and nodes, without the need for extra constituent nodes), a more straightforward mapping to semantic representations (namely, predicate-argument structures) and a more natural support for free word order, making it easier to generalize to different languages and less Anglo-Centric -- a testament to this is the Universal Dependencies project, which offers freely-downloadable dependency treebanks of over 100 languages \citep{ud212}. A more detailed explanation of these advantages can be found in \citep[Chapter 3]{lynnthesis}.

In its most general form, the structure of a sentence $w_1 \ldots w_n$ in dependency grammar can be defined as a directed graph $G = (V,A)$, where

\begin{itemize}
\item the set of nodes is $V = \{ 1 , \ldots , n \}$, each corresponding to a word occurrence in the sentence,
\item $A \subseteq V \times V$ is a set of directed arcs. Each arc of the form $(i, j)$ represents a syntactic dependency from word $w_i$ to word $w_j$, i.e., where $w_i$ is the head of $w_j$ and $w_j$ is the dependent of $w_i$, which can be represented as $w_i \rightarrow w_j$.\footnote{We use the notation where dependencies go from head to dependent as it has become the de facto standard in the last decades, but the reader is warned that definitions where arrows go from dependencies to heads can be found in the literature.}
\end{itemize}

However, syntactic theories typically do not allow arbitrary graphs. The overwhelming majority of them (with a few exceptions such as Word Grammar \citep{hudsonword}) require the structure to be a tree \citep{kubler2009dependency}. Equivalently, the tree constraint can be seen as adding two constraints:
\begin{itemize}
\item Single-head: each node has exactly one head, except for one (the root) which has none.
\item Acyclicity: the graph cannot have cycles (i.e., there cannot be any sequence of one or more dependencies such that we can start at a given word, follow said dependencies in the direction of the arcs, and arrive at the starting word).
\end{itemize}

\begin{figure}[tbp]
\centering
\includegraphics[width=0.9\columnwidth]{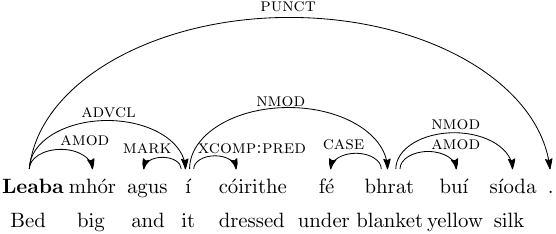}
\caption{A dependency tree in Universal Dependencies format for an Irish sentence (``A large bed dressed in a yellow silk blanket''). The root is shown in boldface.}
\label{fig:A}
\end{figure}

When the graph $G$ satisfies these constraints, we will call it a \emph{dependency tree} for $w_1 \ldots w_n$. An example of a dependency tree for an attested sentence is shown in \cref{fig:A}. Note that, as can be seen in the figure, dependency arcs are labelled with syntactic functions (like \textsc{AMOD} for an adjectival modifier). However, these labels do not affect the constraints defined in this chapter, so we will omit them from now on.

In NLP, dependency grammar is used in dependency parsing, which is the process of automatically finding a dependency tree for a given sentence. In parsing, we can divide most approaches into those where treeness is an integral part of the algorithm design (examples are some transition-based parsers such as arc-standard \citep{nivre-2008-algorithms}, as well as all chart parsers \citep{GomCarWeiCL2011}) and those where the algorithm is suited to parsing graphs, but the tree constraint is enforced explicitly (examples are maximum-spanning-tree-based parsers \citep{mcdonald-etal-2005-non,biaffine}, some transition-based parsers such as Covington \citep{nivre-2008-algorithms} or pointer-network parsers \citep{FerGomNAACL2019}, and sequence-labeling parsers \citep{strzyz-etal-2019-viable}).

In some definitions, an extra node that acts as a dummy root is added to the dependency tree, typically identified with index $0$ and thus located to the left of the nodes representing actual words, although it can also be placed on the right \citep{ballesteros-nivre-2013-squibs}. This is typically just done as a technical convenience so that all words have a head, including the syntactic root which has the dummy root as head. This simplifies file formats, parsing algorithms and sequence-labeling encodings, avoiding the need of special cases for the syntactic root. However, it does sometimes have an impact in the definition of formal constraints on dependency trees, as will be seen in the distinction between projective and planar trees below.

From this point on, we will take the tree constraint for granted and focus on constraints on \emph{which} trees should be allowed. Regardless of the particular syntactic description that one adopts, it has been noted that not all trees are equally plausible in actual language usage. For example, it has long been observed that dependencies seldom cross when drawn above the words \citep{Hays1964,FerGomEstPhysA2018}. More recently, research on dependency distance (the distance of a dependency between words $w_i$ and $w_j$ is the absolute value of the difference $|j-i|$) has demonstrated that dependencies are predominantly short-range even after normalizing distances~\citep{FerGomEstAlePRE2022}. This has motivated researchers, both from engineering and linguistic communities, to implicitly or explicitly pursue the research question of \emph{what formal constraint on dependency trees better characterizes real-life dependency trees}. The interest of this question is at least threefold:
\begin{itemize}
\item From a \emph{linguistic} point of view, a tighter definition of dependency trees, which matches the structures that can actually be found in linguistic analyses of attested language, provides a more accurate linguistic description.
\item From an \emph{engineering} point of view, a more restricted definition of which dependency trees are allowed provides a smaller search space when parsing (finding the adequate dependency tree for a sentence). This means that parsing algorithms can often be made faster (due to fewer possibilities to explore) and less error-prone (for the same reason: fewer possibilities to explore means fewer wrong trees that one could generate instead of the real tree).
\item From a \emph{psycholinguistic} point of view, restrictions on dependency trees can provide insights on human language processing and language evolution. For example, the fact that dependency distances are small can be linked to memory constraints in human processing.
\end{itemize}

The most influential constraint on dependency trees that has been proposed is \emph{projectivity}, which we tackle in the next section. However, it has been seen to be too restrictive, leaving out many attested phenomena in real languages. Therefore, we will discuss more nuanced alternatives.

\section{Projectivity}

\label{sec:projectivity}

Let $T$ be a dependency tree for a sentence $w_1 \ldots w_n$, with $V = \{ 1 , \ldots , n \}$ and $A \subseteq V \times V$ as per the definition above. We first introduce some useful notation that we will need throughout the paper: we define an \emph{interval} as a subset of the form $\{i, i+1, \ldots , j-1, j\} \subseteq V$, written in shorthand as $[i,j]$, i.e., a set containing all nodes spanning from $i$ to $j$. We write $i \xrightarrow{*} j$ to indicate that there is a (possibly empty) path from $i$ to $j$, and we will call $j$ a \emph{descendant} of $i$.\footnote{The descendant relationship is reflexive as well as transitive, i.e., $i$ is a descendant of itself.} Then, the \emph{projection} of node $i$ is the set $\{j : i \xrightarrow{*} j\}$, i.e., the set formed by node $i$ and all of its descendants. We say that a dependency $i \rightarrow j$ \emph{covers} node $m$ if $\min(i,j)<m<\max(i,j)$. Finally, we say that dependencies $i \rightarrow j$ and $k \rightarrow l$ \emph{cross} if either $\min(i,j)<\min(k,l)<\max(i,j)<\max(k,l)$ or $\min(k,l)<\min(i,j)<\max(k,l)<\max(i,j)$. These two latter notions manifest graphically as the arrow of dependency $i \rightarrow j$ passing above node $m$ and crossing the arrow of dependency $k \rightarrow l$ when both are drawn above the words.

\begin{figure*}[tbp]
    \centering
    \includegraphics[width=1.5\columnwidth]{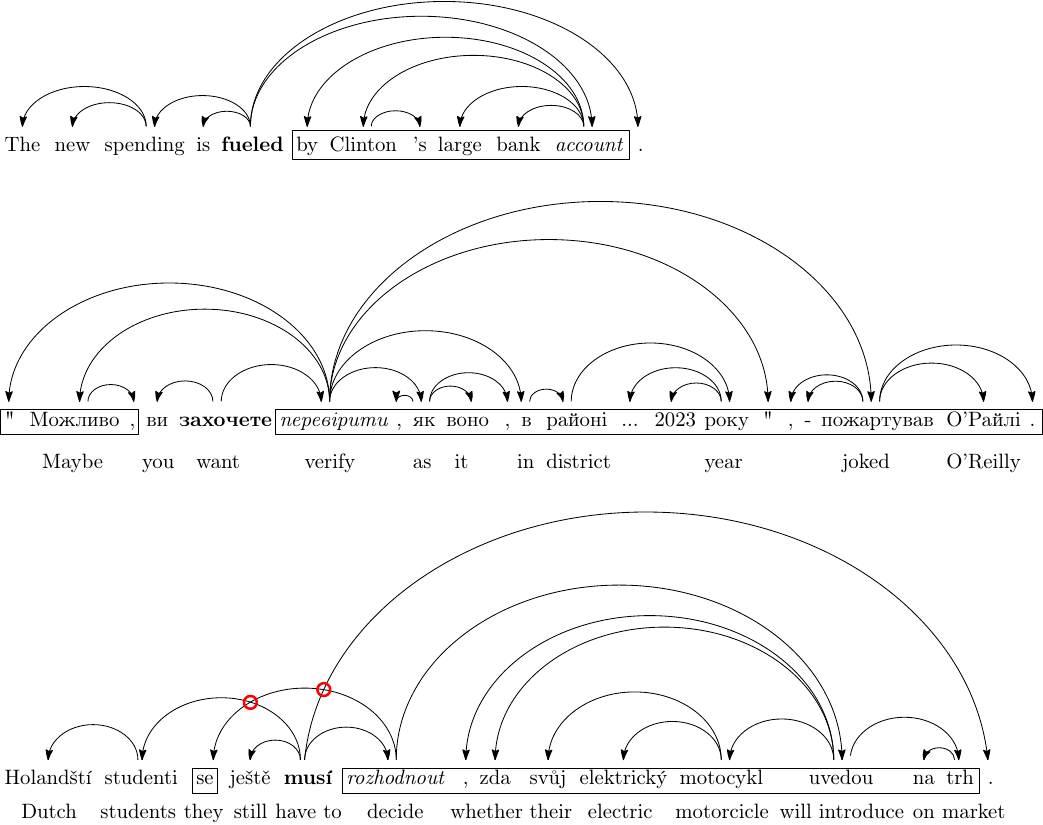}
    \caption{Projectivity and 1-planarity. For each tree, the root is shown in boldface, and we show the projection of the word highlighted in italics. The top tree is projective and planar (the continuous projection of \emph{account} is highlighted). The tree in the middle is planar, but not projective (the projection of \cyrins{\emph{перевірити}} is discontinuous). The bottom tree is neither projective nor planar (it has a discontinuous projection and arc crossings, marked with red circles). The second sentence is taken from a Ukrainian UD treebank (version 2.6); it can be translated as ``"You might want to check out what it's like around ... 2023," O'Reilly joked''. The third sentence is taken from a Czech parallel UD (PUD) treebank (version 2.6); it can be translated as ``The Dutch students have yet to decide if they will be commercializing their electric motorcycle''.}
    \label{fig:B}
\end{figure*}

With this, projectivity can be defined in various equivalent ways. The simplest is probably the following: tree $T$ is \emph{projective} if, and only if, the projection of every node in $T$ is an interval. This corresponds to the yield (set of descendants) of each node being a continuous piece of the input sentence. Thus, the top tree in \cref{fig:B} is projective (an example of continuous projection is shown), while the middle and bottom trees are not (as they have at least one node whose projection is discontinuous, also highlighted in the figure). Equivalently, one can say that $T$ is projective if, and only if, for every one of its dependencies with head $h$ and dependent $d$, every node covered by the dependency is a descendant of $h$. A dependency not satisfying this condition (i.e., where there is some intervening material between head and dependent that does not descend from the head) is called a non-projective dependency. In \cref{fig:B}, the dependencies from \cyrins{\emph{перевірити}} to \cyrins{\emph{Можливо}} and from \emph{rozhodnout} to \emph{se} are non-projective.

Projectivity is also commonly associated with the absence of crossing arcs, but to understand that relationship, we need to define a related restriction: planarity, also called 1-planarity.\footnote{Planarity is the dominant term in the dependency parsing literature, but we will use the term 1-planarity to prevent confusion with planarity in graph theory (which is \emph{not} an equivalent term) as well as to make the link to the related concept of 2-planarity more natural.} A dependency tree is \emph{1-planar} if it contains no crossing dependencies (equivalently, it can also be called a non-crossing or pagenumber-1 tree). Under the assumption that all trees have a dummy root at position 0, projectivity and 1-planarity are equivalent, i.e., projective trees are those that do not have crossing dependencies. However, if a dummy root is not used, the set of 1-planar trees is a strict superset of the set of projective trees. The only trees that are 1-planar but not projective are those where the syntactic root is covered by a dependency, which have been found to be relatively scarce in practice \citep{gomez-rodriguez-nivre-2013-divisible}. This observation allows us to provide another alternative definition of projectivity: a tree is projective if it is 1-planar and no arc covers the root. The middle sentence shown in \cref{fig:B} has a tree that is 1-planar, but not projective (the root, in boldface, is covered by an arc), whereas the bottom tree is not 1-planar because it has two dependency crossings.

Projectivity has traditionally been, and still is, the most well-known and used constraint on dependency trees both for linguists and for NLP practitioners. From the linguistics point of view, various linguistic theories, especially focused on languages like English \citep{sleator93,hudson07} or Japanese \citep{Tanaka97,KyotoCorpus}, have traditionally assumed that crossing dependencies do not exist. According to \citet{Ninio2017}, projectivity, ``the mathematical code of syntax'', is ``pervasive, ubiquitous, and probably universal''. Indeed, it has been statistically shown that syntactic analyses of human language exhibit a trend for projectivity, as crossing dependencies are considerably scarcer than would be expected \citep{FerGomEstPhysA2018}. However, the claims that projectivity is ubiquitous and universal, or that non-projectivity is unneeded to describe syntax, do not fit well with flexible-word-order languages such as German, where non-projectivity is very common in treebanks (cf. data from \citep{GomCL2016,Decatur22}). The issue can be even more marked if we explore non-standard genres and styles of language: for example, in a treebank of Ancient Greek containing poetry, the ratio of non-projective sentences has been reported to be almost 75\% \citep{mambrini-passarotti-2013-non}, so it is safe to say that the claim that projectivity is universal seems untenable in light of the data. More defensible is the position that the syntax of some particular languages, like English or Japanese, could be fully projective, as some syntactic theories have defined them to be. However, this still clashes with the fact that other theories, such as Universal Dependencies or Surface Syntactic Universal Dependencies, do find non-projectivity in English due to phenomena like wh-movement or topicalization, even if the prevalence is low (see \citep{Decatur22} for statistics). Advocates of the projectivity of these languages could say that the theories that cast them as fully projective are ``elegant theoretical solutions'' \citep{Ninio2017}. However, such theories could also be seen as a form of observer bias: if projectivity is seen as desirable, be it to ease parsing or for aesthetical considerations \citep{FerGomComplexity2016}, theory designers and annotators might try to find a way to analyze every sentence as projective, even if this is not necessarily the most coherent, parsimonious or cognitively realistic way to do so. 

Be that as it may, the projectivity constraint has become highly influential in natural language processing, largely because of the field's disproportionate focus on English \citep{bender2019rule}, a language in which non-projectivity is infrequent, or even nonexistent in specific theories and treebanks. Projectivity has clear advantages for syntactic parsing, as it allows for simpler and faster parsers in almost all paradigms. In transition-based parsing, where the syntax tree is built by a non-deterministic state machine, projective parsing can be done by simple linear-time algorithms that operate on a stack and have a branching factor of three transitions, while non-projective support needs adding more complicated data structures and extra transitions \citep{nivre-2008-algorithms,gomez-rodriguez-nivre-2013-divisible}. In parsing as sequence labeling, where the tree is encoded as a sequence of one label per word obtained by a tagger, projectivity enables simpler and more compact encodings \citep{amini-etal-2023-hexatagging,gomez-rodriguez-etal-2023-4}. The parsing paradigm where the advantages of projectivity are less marked in practice is probably graph-based parsing, where the syntactic analysis is obtained by scoring and combining tree fragments, since there is a family of parsing algorithms based on maximum spanning trees which is fast, efficient and rather popular, even for English \citep{biaffine}, and they support full non-projectivity. Even in this paradigm, though, projectivity can be advantageous as it can enable the use of more detailed features \citep{fonseca-martins-2020-revisiting}.

For these reasons, the overwhelming majority of parsers used to be projective two decades ago; and even now, in spite of hardware advances and better machine learning techniques that make non-projective approaches feasible, projective parsers are still common. For example, current state-of-the-art results on the well-known English Penn Treebank benchmark are held by a very recent projective sequence-labeling-based parser \citep{amini-etal-2023-hexatagging}. Also worth noting is that this parser even achieves state-of-the-art performance in some languages with nontrivial amounts of non-projectivity, like Czech, via pseudo-projectivity \citep{nivre-nilsson-2005-pseudo}, a transformation that can convert non-projective to projective trees (albeit adding complexity to their arc labels) and back. For other languages like German, however, this approach is clearly outperformed by parsers that support non-projectivity natively, without the need of such a transformation. At any rate, the availability of pseudo-projectivity means that, even in multilingual environments, projective parsers still have a relevant place and should not be neglected, especially when simplicity and efficiency is desired. 

\section{Mild Non-Projectivity}

As outlined in the previous section, while some linguistic theories assume that language is best modeled by fully projective analyses (often framed as the claim that language itself is projective), this does not fit well with observations, especially (but not exclusively) in flexible-word-order languages. Thus, a realistic grammatical theory should allow dependency arcs to cross. However, assuming that human languages can have unrestricted non-projectivity (i.e., any combination of crossing arcs) is clearly overkill from several standpoints. From a linguistic point of view, non-projectivity has been found to be scarce, with real syntactic trees showing much fewer dependency crossings than one would expect by chance \citep{FerGomEstPhysA2018}, so it would not make sense to assume that any arrangement of dependents is grammatically plausible. From a psycholinguistic point of view, limitations in human working memory mean that dependencies generally need to be short for sentences to be understandable, and this in turn implies fewer crossing dependencies \citep{FerGomEstAlePRE2022,FerGomComplexity2016}. Finally, from an engineering point of view, supporting full non-projectivity comes with an associated computational cost, e.g.\  quadratic rather than linear transition-based parsers \citep{nivre-2008-algorithms} or graph-based parsing being outright intractable except with very simple feature models \citep{mcdonald-satta-2007-complexity}.

For these reasons, researchers have tried to find \emph{mildly non-projective} sets of dependency trees, i.e., sets that are less restrictive than the set of projective trees, but more restrictive than arbitrary non-projectivity. A desirable mildly non-projective set of trees should strike a careful balance: on the one hand, it should be large enough to support the non-projective phenomena that actually appear in analyses of attested language usage (in order to have good coverage of real sentences, be it for analysis by linguists or for natural language processing applications). On the other hand, it should be restrictive enough, although the reason for this can change depending on purpose and in turn affect which sets are better for a particular scenario: from a linguistics or psycholinguistics point of view, the main reason to restrict non-projectivity is to prevent the grammar from covering non-realistic syntactic structures, thus providing a more accurate linguistic restriction. From an engineering point of view, the interest is in efficiency, and a mildly non-projective set will be more useful if there is an efficient parsing algorithm that supports it. These two goals are not always linked, i.e., some mildly non-projective sets were defined because they were found to be linguistically interesting but parsers for them are not especially efficient (or even do not exist at all), and vice versa.

We now proceed to describe the main mildly non-projective sets of trees that have been defined in the literature. It is worth noting that coverage and parsing complexity statistics of many of these sets can be found in \citep{GomCL2016}, and all coverage figures and comparisons provided below are taken from that paper. However, this source does not cover sets defined after its publication in 2016.

\subsection{Sets related to well-nestedness and gaps}

\begin{figure}[tbp]
\begin{center}
\includegraphics[width=0.8\columnwidth]{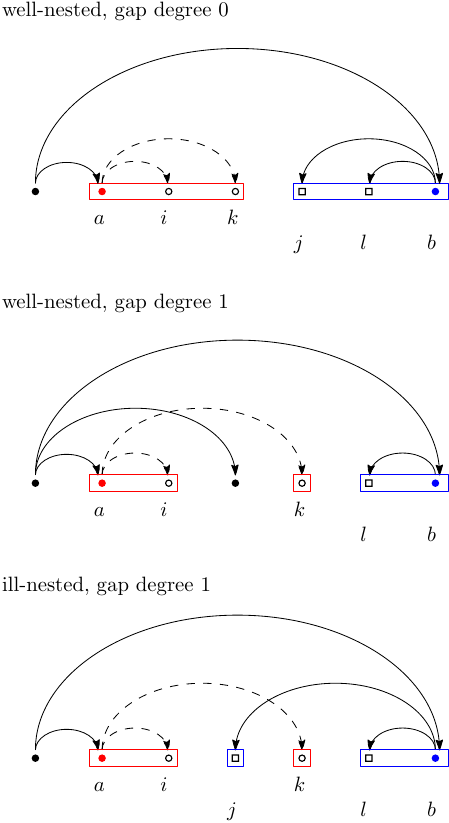}
\end{center}
\caption{Well-nestedness and gap degree. The top tree is well-nested and has gap degree 0 (two example projections, which are continuous and do not interleave, are shown in red and blue). The middle tree has gap degree 1 (as the projection of $a$ has a gap) but it is still well-nested. The bottom tree has gap degree 1 and is ill-nested, as the projections of $a$ and $b$ interleave.}
\label{fig:C}
\end{figure}

A dependency tree is said to be \emph{ill-nested} \citep{bodirsky05} if it contains two nodes whose projections \emph{interleave}, i.e., if there are nodes $a,b$ and $i<j<k<l$ such that $i$ and $k$ are in the projection of $a$ but not of $b$, whereas $j$ and $l$ are in the projection of $b$ and not of $a$. A tree that is not ill-nested is said to be \emph{well-nested}. In \cref{fig:C}, the bottom tree is ill-nested (the interleaving projections of nodes $a$ and $b$ are highlighted) whereas the other two are well-nested. 

Ill-nestedness has been shown to be very rare, although not entirely absent, in real syntax trees \citep{chen-main-joshi-2010-unavoidable}; and well-nestedness, together with other properties, has been used to define several mildly non-projective sets of interest.

Probably the best known among these is the set of \emph{well-nested trees with gap degree $\le$ k} \citep{bodirsky05}, as it was one of the first mildly non-projective sets to be studied, while being (still to this day) one of the few that has both dedicated parsers and a relation to preexisting grammatical theories. This set is defined based on the concept of \emph{gap degree}: the gap degree of a node $x$ in a dependency tree is the smallest integer $k$ for which the projection of $x$ can be expressed as the union of $k+1$ intervals. Intuitively, this corresponds to the number of gaps in the projection, i.e., a node with projection $\{3,4,5\} = [3,5]$ has gap degree zero, whereas a node with projection $\{1,2,3,6,7,8,12,13,14\} = [1,3] \cup [6,8] \cup [12,14]$ has gap degree two, as its projection has two gaps (corresponding to the intervals $[4,5]$ and $[9,11]$). From this, the gap degree of a dependency tree is defined as the highest gap degree among its nodes. Thus, trees with gap degree zero correspond to projective trees (as the projection of each of its nodes is an interval, which matches the first definition of projectivity provided in \cref{sec:projectivity}). By bounding the gap degree to values higher than 0, we can define successively milder constraints on non-projectivity. In \cref{fig:C}, the top tree has gap degree $0$ while the other two have gap degree $1$ (notice the gap in the projection of node $a$ in each of them).

Well-nestedness and gap degree are of limited practical interest in isolation (in the sense that there is no known parser whose coverage is well-nested trees as a whole, and the same applies to bounded gap degree for $k>0$). However, they become relevant when intersected. Namely, the set of trees that are well-nested and whose gap degree is bounded by a constant $k$ is $WG_k$, the set of \emph{well-nested trees with gap degree $\le k$}. While $WG_0$ corresponds to projective trees, $WG_k$ sets for $k>0$ are mildly non-projective sets with the following interesting properties:
\begin{itemize}
\item Under annotation criteria that use content words as heads (Stanford dependencies \citep{de-marneffe-manning-2008-stanford}, similar to the dependency annotation that is now most prevalent, Universal Dependencies) $WG_1$ was shown to cover over $95\%$ of the dependency trees in a variety of treebanks, and this coverage is pushed to around $98\%$ with $WG_2$. The figures are even higher ($\sim 98\%$ for $WG_1$, $>99\%$ for $WG_2$) under Prague dependencies \citep{hajic-etal-2020-prague}, a syntactic theory that prefers function words as heads, in a similar way as Surface-Syntactic Universal Dependencies (SUD) \citep{gerdes-etal-2018-sud}. 
\item $WG_k$ is parsable with a dynamic programming algorithm that finds the highest-scoring tree in time $O(n^{5+2k})$ \citep{GomCarWeiCL2011}.
\item The $WG_k$ sets have a direct relation with (initially constituency-based) grammatical formalisms: $WG_1$ trees correspond to the dependency trees induced by tree-adjoining grammars \citep{Joshi1997}, and $WG_k$ trees are induced by coupled context-free grammars \citep{kuhlmann2007dependency}.
\end{itemize}

\begin{figure}[tbp]
\begin{center}
\includegraphics[width=1.0\columnwidth]{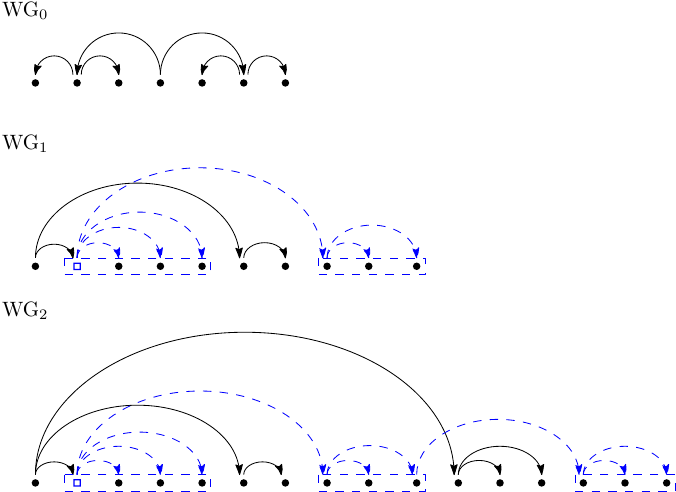}
\end{center}
\caption{The $WG_k$ sets. The top tree is in $WG_0$, the middle tree is not in $WG_0$ but it is in $WG_1$ (a projection with gap degree 1, but which does not interleave any other projection, is highlighted), and the bottom tree is not in $WG_0$ or $WG_1$ but it is in $WG_2$ (a projection with gap degree 2, but which does not interleave any other projection, is highlighted).}
\label{fig:D}
\end{figure}

Note that, in \cref{fig:C}, the top tree is in $WG_0$; the middle tree is not in $WG_0$ but it is in $WG_1$; and the bottom tree is not in $WG_k$ for any value of $k$, since it is ill-nested. Complementarily, \cref{fig:D} shows trees from $WG_0$, $WG_1$ and $WG_2$, respectively.

The combination of high coverage, parsability and linguistic relevance has fueled a sustained interest in these sets (especially $WG_1$, with its more manageable complexity) over the years, see e.g.\ parser implementations by \citet{corro-etal-2016-dependency,corro-2020-span}. In fact, some authors have even referred to $WG_1$ as ``\emph{the} class of mildly non-projective trees'' \citep{pitler-etal-2012-dynamic}, which gives an idea of how influential this set has been. In addition, various restrictions and extensions of the $WG_k$ sets have been defined, which we now describe:
\begin{itemize}
\item \citet{GomCarWeiCL2011} define an extension of $WG_k$ called $MG_k$, the set of \emph{mildly ill-nested dependency trees with gap degree $k$}. A tree is in $MG_k$ if it can be converted to a binary tree (by decomposing nodes to reduce out-degree) in such a way that said binary tree has gap degree $k$. Each set $MG_k$ is a superset of $WG_k$, and practical coverage gains are substantial. However, the complexity of parsing $MG_k$ trees is $O(n^{4+3k})$ (cf. $O(n^{5+2k})$ for $WG_k$). Thus, this extension is especially interesting for gap degree $1$, where it comes at no extra complexity cost since both $WG_1$ and $MG_1$ are parsable in $O(n^7)$.
\item In the opposite direction, \citet{pitler-etal-2012-dynamic} add an extra constraint to $WG_1$ based on the concept of gap inheritance. A child node is said to ``inherit'' the gap in the projection of its parent's node if it has descendants on both sides of the gap. With this, they define \emph{Mild+1-Inherit ($M1I$) trees} (at most one child can inherit the parent's gap) and \emph{gap-minding (M0I) trees} (no child can inherit the parent's gap). These sets can be parsed more efficiently than $WG_1$ trees (having parsers that run in $O(n^6)$ and $O(n^5)$, respectively) and in the case of Mild+1-Inherit trees this comes at almost no cost in empirical coverage, albeit this is no longer the case with gap-minding trees.
\item \citet{satta-kuhlmann-2013-efficient} add another extra constraint to $WG_1$, called the head-split property: if the gap in the projection of a node contains the node's head, then it must also contain the gap in the projection of said head. This constraint can be added either directly to $WG_1$, or to the set of $M1I$ trees (i.e., enforcing the intersection of both the head-split and Mild+1-Inherit constraints), and its effect is reducing the complexity by a factor of $n$ with negligible loss of coverage. Thus, head-split $WG_1$ trees can be parsed in $O(n^6)$, while head-split $M1I$ trees can be parsed in $O(n^5)$ and have higher coverage than gap-minding trees.
\item A third, alternative restriction to bring the complexity of parsing a subset of $WG_1$ trees to $O(n^6)$ is shown by \citet{corro-2020-span}: one can implement a variant of the $O(n^7)$ $WG_1$ parser of \citep{GomCarWeiCL2011}, but remove the costliest deduction rule; which is the complexity bottleneck. Corro shows that this has little effect in coverage while reducing time complexity by a factor of $n$. The same trick can be applied to mildly ill-nested parsers. A drawback with respect to the previously-defined restrictions is that this one has a purely operational definition (i.e., it is defined by what a specific variant of a parser can parse). On the other hand, it has the benefit that it is easy to have the same parser operate with or without this restriction. Faster to parse (but narrower in coverage) sets can also be defined by removing more rules from the parser.
\end{itemize}

\subsection{Sets related to page number}

\begin{figure}[tbp]
\begin{center}
\includegraphics[width=1.0\columnwidth]{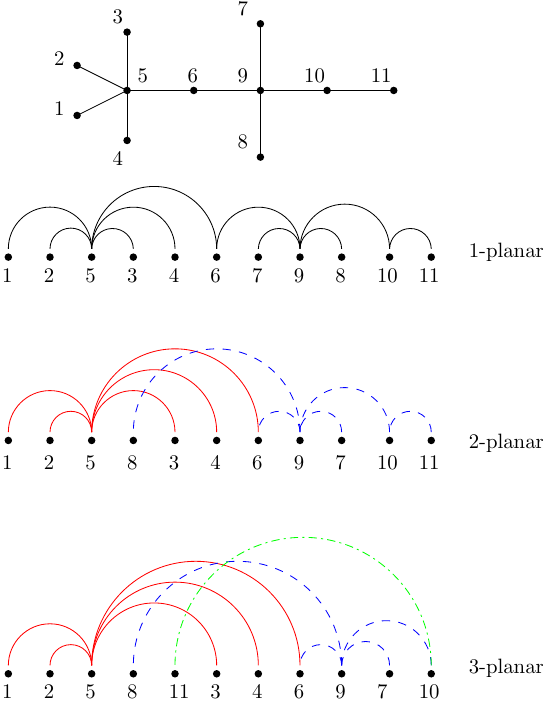}
\end{center}
\caption{Multiplanarity. The 1-planar tree has no dependency crossings. In the 2-planar tree, we can split the arcs into two planes (associated here with colors) such that arcs within the same plane do not cross. The last tree is not 2-planar, so it is not possible to do this with two colors, but it is 3-planar, since it is possible with three colors, as shown. Note that the three dependency trees are \emph{linear arrangements} of the tree shown above: they have the same structure, but the order of the words (represented by numbers) changes. Indeed, projectivity, planarity, and all the other properties discussed here can be seen as properties of word order. In fact, every dependency tree can be reordered into a projective tree, and some parsing strategies have exploited this \citep{nivre-2009-non}.}
\label{fig:F}
\end{figure}

Another family of mildly non-projective sets of trees, unrelated to the previous ones, comes from the concept of \emph{page number}, which has often been called \emph{multiplanarity} in the dependency parsing literature  \citep{yli03}. A graph (or in particular, a dependency tree) has \emph{page number $k$} (or equivalently, is \emph{$k$-planar}) if its arcs can be partitioned into $k$ subsets (called pages or planes) such that arcs assigned to the same subset do not cross. It is often useful to visualize each subset with a different color: for example, a 2-planar dependency tree is one where the arcs can be drawn above the words in either of two colors (say, red and blue) in such a way that arcs of the same color do not cross. \cref{fig:F} shows examples of 1-, 2- and 3-planar trees where the arcs have been colored as a function of the subset to which they have been assigned. Notice that there may be more than one assignment of arcs to subsets that satisfy the property of $k$-planarity.

\begin{figure}[tbp]
\begin{center}
\includegraphics[width=1.0\columnwidth]{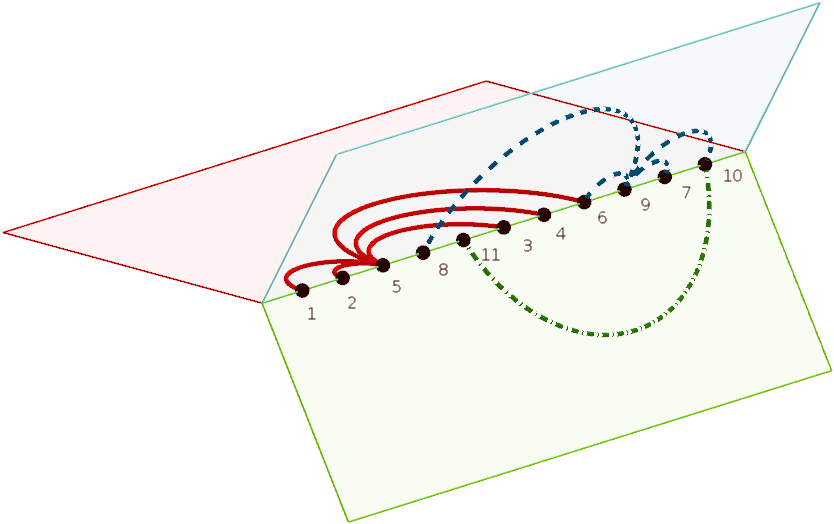}
\end{center}
\caption{A ``book embedding'' 3D view of the bottom tree in \cref{fig:F}. The tree is 3-planar because its arcs can be drawn on three pages of a book, such that arcs within the same page do not cross.}
\label{fig:F2}
\end{figure}

The page-related terminology comes from imagining each subset being drawn into a different page of a book, where nodes occupy the spine of the book. \cref{fig:F2} shows the ``book embedding'' view of the same 3-planar tree where the arcs are embedded into pages of a book.

When $k=1$, the concept of $k$-planarity corresponds to the $1$-planarity that we already defined in \cref{sec:projectivity} and, as mentioned there, is only a very mild extension of projectivity (or even equivalent to it, when a dummy root node is used). However, if we increase $k$ even to only 2, we obtain a very wide-coverage extension: in the same sets of treebanks quoted in the previous section, around $99.5\%$ of the trees are $2$-planar, which is better coverage than $WG_1$ and $MG_1$, and comparable to $WG_2$.

With respect to parsing, no polynomial parser to find the highest-scoring tree (like the ones discussed in the previous section) is known for $k$-planar trees, but there is a linear-time transition-based parser \citep{gomez-rodriguez-nivre-2013-divisible}. This is a greedy algorithm that does not guarantee optimality, but it has been shown to produce reasonable empirical accuracy for $k=2$. This arguably makes 2-planarity a more useful restriction for practitioners than the ones covered in the previous section: while exact inference (i.e., guaranteeing the highest-scoring tree) is an interesting theoretical property, a $O(n)$ approximate parser with good empirical accuracy is better for most uses than a $O(n^5)$ (or more) exact-inference parser. However, on the other hand, it can be argued that 2-planarity is less linguistically interesting, lacking the connections to formal grammar theory of $WG_k$ and its variants.

\begin{figure}[tbp]
\begin{center}
\includegraphics[width=1.0\columnwidth]{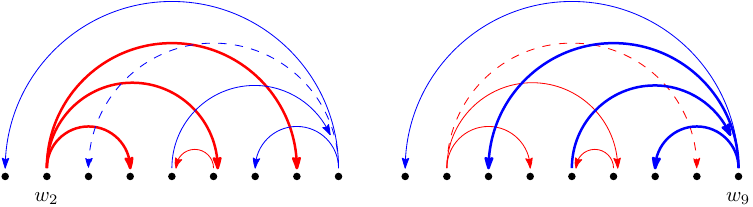}
\end{center}
\caption{Two views of the same 1-Endpoint-Crossing tree. The left view highlights the arcs crossing one given arc (dotted), showing that they have the node $w_2$ in common. The right view focuses on a different arc (dotted), such that all its crossing arcs have $w_9$ in common. Colors represent planes, showing that the tree is also 2-planar, like all 1-Endpoint-Crossing trees.}
\label{fig:G}
\end{figure}

In this respect, a subset of 2-planar trees has been defined that has an exact inference dynamic programming parser: the set of \emph{1-Endpoint-Crossing} trees \citep{pitler-etal-2013-finding}, which is the set of trees such that all edges crossing one given edge have a common vertex, and can be parsed in $O(n^4)$, i.e., faster than the sets seen in the previous section, which facilitated a practical implementation \citep{pitler-2014-crossing}. In addition, the coverage of this set is better than that of e.g.\ $WG_1$ and it is the largest among known mildly non-projective sets that are parsable with exact inference in $O(n^4)$; and the 1-Endpoint-Crossing constraint also has some linguistic and psycholinguistic backing, discussed in \citep{pitler-etal-2013-finding}. 
A 1-Endpoint-Crossing tree can be seen in Figure \ref{fig:G}, showing how all arcs crossing a given arc are incident to a common node.

In turn, a more restricted subset of 1-Endpoint-Crossing trees are \emph{$k$-Crossing-Interval} trees, which are defined by an elaborate constraint on how crossing dependencies are arranged with respect to intervals of the input string. These, like general 2-planar trees, can be parsed by a linear-time transition system \citep{pitler-mcdonald-2015-linear}.

Finally, it is worth noting that restrictions on non-projectivity based on the concept of $k$-planarity have also found relevance in sequence labeling parsing, a parsing paradigm where syntax trees are represented by sequences of one label per word. In particular, several bracketing-based encodings to perform sequence labeling parsing \citep{StrVilGomCOLING2020,gomez-rodriguez-etal-2023-4} operate on an extension of $k$-planar trees where arcs in opposite direction are allowed to cross (i.e., graphs that can be partitioned into $k$ subsets so that arcs \emph{in the same direction} and assigned to the same subset do not cross). 

\subsection{Operationally-defined sets}

Other mildly non-projective sets of dependency trees have been defined operationally, i.e., these are sets whose definition is ``the set of dependency trees covered by the given parser(s) X'', and such that no simple, declarative definition is known that does not involve a parsing algorithm. This arguably makes these sets less interesting from a theoretical standpoint, but they can still be relevant from an empirical standpoint if their associated parsers have a good coverage/efficiency tradeoff.

A first family of such sets is defined by the Attardi parser \citep{attardi-2006-experiments} and its modified version by \citet{cohen-etal-2011-exact}, which is a variant of the arc-standard projective parser \citep{nivre-2008-algorithms} which adds transitions that create arcs that ``skip'' over nodes on the stack (in contrast to arc-standard, which only links adjacent nodes, hence allowing no crossing arcs), while keeping $O(n)$ complexity. By allowing the transitions to skip over $k$ nodes, we obtain the trees of Attardi degree $k+1$, or $AD_{k+1}$. Much like with $k$-planarity, interest has especially focused on degree 2, as it provides a rather good coverage (in between that of $WG_1$ and 2-planarity) while not complicating the parser too much, as larger degrees require adding more transitions and this makes effective learning more difficult for the model. In addition to the original transition-based parser, a dynamic programming algorithm allowing exact inference for parsing $AD_k$ trees was defined by \citet{cohen-etal-2011-exact} with complexity $O(n^{3k+1})$. Later on, \citet{ShiGomLeeNAACL2018} defined several variants of this algorithm (in its version for degree $2$) that reduce the complexity to $O(n^6)$ while increasing coverage. The sets covered by these algorithm variants are also operationally defined.

\begin{figure}[tbp]
\begin{center}
\includegraphics[width=1.0\columnwidth]{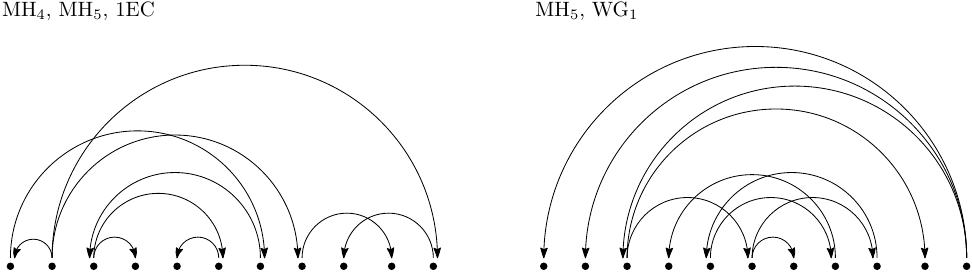}
\end{center}
\caption{The $MH_k$ sets. The left tree is in $MH_4$ (and is also 1-Endpoint-Crossing, but it is not well-nested). The right tree is not in $MH_4$, but it is in $MH_5$ (and well-nested, but not 1-Endpoint-Crossing). Note that while well-nestedness and the 1-Endpoint-Crossing property can be easily inferred from the graphical representation, it is not know how to do this for the operationally-defined $MH_k$ sets.}
\label{fig:I}
\end{figure}

On the other hand, in a separate research line from the Attardi parser, \citet{GomCarWeiCL2011} defined a dynamic programming parser called $MH_k$, meaning ``multi-headed with $k$ heads per item'', which is a generalization of the dynamic programming version of the arc-hybrid projective parser \citep{kuhlmann-etal-2011-dynamic}. In particular, from said parser one can obtain a projective dynamic programming algorithm based on combining items that represent forests with at most three roots ($MH_3$). By increasing the number of allowed heads beyond 3, one can accept successively larger sets of non-projective structures. The complexity of the $MH_k$ parser is $O(n^k)$, and while for $k=4$ its coverage is slightly worse than that of 1-Endpoint-Crossing trees, for $k>4$ it is the set with the best tradeoff between coverage and complexity of dynamic programming algorithms \citep{GomCL2016}, thus making it practically relevant even if it lacks a non-operational definition. \cref{fig:I} shows a tree in $MH_4$ and another that is not in $MH_4$, but belongs to $MH_5$.

$MH_k$ trees are related to Attardi trees in the sense that $MH_{d+2} \subseteq AD_d$. But a closer relation was later found by \citet{gomez-rodriguez-etal-2018-global} who, apart from providing a practical implementation of $MH_4$, also converted it back to a transition-based parser, obtaining an Attardi-like transition-based parser which covers $MH_4$ trees in linear time.

\section{Conclusion}

We have explored the complex landscape of formal constraints on dependency syntax. The journey began with an in-depth examination of projectivity, a foundational concept in dependency grammar, revealing its limitations in fully capturing human syntax, especially in flexible-word-order languages. This limitation paved the way for an exploration of mildly non-projective trees, which offer a more realistic middle ground between the rigidity of projectivity and the over-generality of unrestricted dependency structures.

\begin{figure*}[tbp]
\begin{center}
\includegraphics[width=1.5\columnwidth]{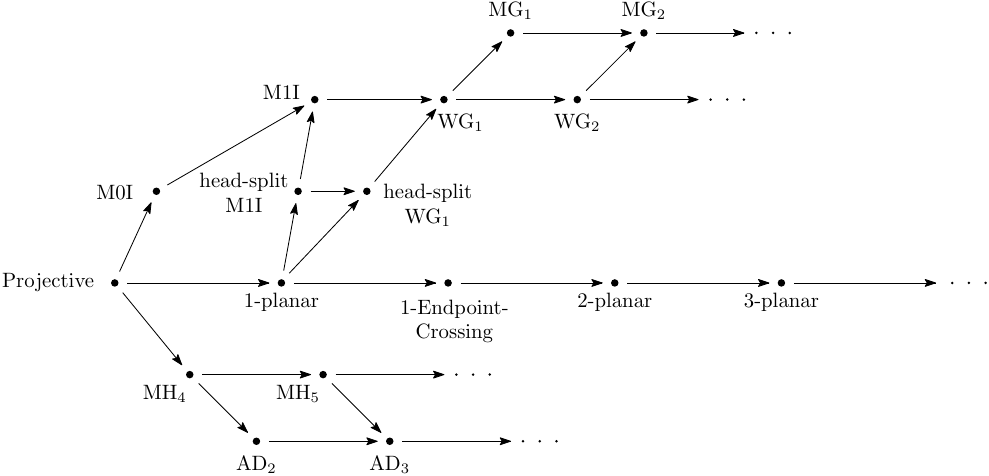}
\end{center}
\caption{An inclusion diagram with the known inclusions between the major classes of dependency trees that we have discussed. A path from set $A$ to $B$ in the graph means that $A \subseteq B$.}
\label{fig:J}
\end{figure*}

The variety of proposals for mildly non-projective sets of trees, which we have divided into three broad groups according to their origin, underscores the diversity and complexity inherent in natural languages. There is no one-size-fits-all approach to constraining non-projectivity, as some alternatives may be better for an engineering perspective (yielding practical, efficient parsers) while others have more linguistic or psycholinguistic interest. \cref{fig:J} shows an inclusion diagram with the relations between the major classes of trees discussed here -- as can be seen, there is no single hierarchy where each class is contained in the next (like the well-known Chomsky hierarchy), but a more complex panorama of interrelated and independent hierarchies.

This area of research has several open avenues for future work. One area is the development of parsing algorithms that keep pushing the trade-off between non-projective coverage, on the one hand, and efficiency and simplicity, on the other. A more underexplored direction is the study of the psycholinguistic implications of these syntactic structures and their relation to phenomena like dependency length minimization, potentially leading to a deeper understanding of the cognitive processes underpinning language comprehension and production \citep{GomChrFerGlotto2022}. For example, one might ask: to what extent do constraints like $WG_1$, $1EC$ or $MH_4$ have such a good coverage of attested syntax due to their definitional properties (e.g. the notion of well-nestedness); and to what extent is their coverage a side effect of these constraints implicitly minimizing metrics like dependency length or number of crossings? Ultimately, a deeper understanding would require modeling the interplay between these formal syntactic constraints and other crucial information sources used by humans, such as semantic plausibility and world knowledge.

\begin{ack}[Acknowledgments]{}
CGR has received funding from Xunta de Galicia (ED431C 2020/11), and Centro de Investigación de Galicia ``CITIC'', funded by the Xunta de Galicia through the collaboration agreement between the Consellería de Cultura, Educación, Formación Profesional e Universidades and the Galician universities for the reinforcement of the research centres of the Galician University System (CIGUS). LAP is supported by a recognition 2021SGR-Cat (01266 LQMC) from AGAUR (Generalitat de Catalunya) and the grants AGRUPS-2022 and AGRUPS-2023 from Universitat Politècnica de Catalunya, and by Secretaria d'Universitats i Recerca de la Generalitat de Catalunya and the Social European Fund and a Ph.D. contract extension funded by Banco Santander.
\end{ack}

\seealso{article title article title}

\bibliographystyle{APA-Style}
\bibliography{reference}

\end{document}